\pdfoutput=1

\documentclass[11pt]{article}

\usepackage{acl}

\usepackage{times}
\usepackage{latexsym}

\usepackage[T1]{fontenc}

\usepackage[utf8]{inputenc}

\usepackage{microtype}

\usepackage[pdftex]{graphicx}
\graphicspath{ {figures/} }
\usepackage{wrapfig}

\usepackage{hyperref}       
\usepackage{url}            
\usepackage{booktabs}       
\usepackage{amsfonts}       
\usepackage{nicefrac}       
\usepackage{xcolor}         
\usepackage{amsmath}

\usepackage{caption}
\usepackage{subcaption}
\usepackage{todonotes}
\usepackage{xcolor}
\usepackage[normalem]{ulem}
\usepackage{enumitem}
\usepackage{subcaption}
\newcommand{\sys}{\textsc{DiSCoMaT}}
\newcommand\FramedBox[3]{%
  \setlength\fboxsep{0pt}
  \fbox{\parbox[t][#1][c]{#2}{#3}}}
\title{\textsc{DiSCoMaT}: Distantly Supervised Composition Extraction \\from Tables in Materials Science Articles}

\author{
    Tanishq Gupta$^{1}$, Mohd Zaki$^{2}$, Devanshi Khatsuriya$^{3}$, Kausik Hira$^{4}$, \\
    \hspace*{0.5cm} \textbf{N. M. Anoop Krishnan}$^{4,2}$, \textbf{Mausam}$^{4,3}$\\
    $^{1}$Department of Mathematics, $^{2}$Department of Civil Engineering \\
    $^{3}$Department of Computer Science and Engineering, $^{4}$Yardi School of Artificial Intelligence \\
    Indian Institute of Technology Delhi \\
    \small{\texttt{\{tanishqg2406, mohdzaki1995, devanshikhatsuriya18, kausikhira\}@gmail.com}} \\
    \small{\texttt{\{krishnan, mausam\}@iitd.ac.in}} \\
}
\begin{document}
\makeatletter\acl@finalcopytrue
\maketitle


\begin{abstract}

A crucial component in the curation of KB for a scientific domain (e.g., materials science, foods \& nutrition, fuels) is information extraction from tables in the domain's published research articles. To facilitate research in this direction, we define a novel NLP task of extracting compositions of materials (e.g., glasses) from tables in materials science papers. The task involves solving several challenges in concert, such as tables that mention compositions have highly varying structures;  text in captions and full paper needs to be incorporated along with data in tables; and regular languages for numbers,  chemical compounds and composition expressions must be integrated into the model. 

We release a training dataset comprising 4,408 distantly supervised tables, along with 1,475 manually annotated dev and test tables. We also present \sys, a strong baseline that combines multiple graph neural networks with several task-specific regular expressions, features, and constraints. We show that \sys{} outperforms recent table processing architectures by significant margins. 
We release our \href{https://github.com/M3RG-IITD/DiSCoMaT}{code and data} for further research on this challenging IE task from scientific tables.

\end{abstract}

\section{Introduction}
\label{sec:intro}

Advanced knowledge of a science or engineering domain is typically found in domain-specific research papers. Information extraction (IE) from scientific articles develops ML methods to automatically extract this knowledge for curating large-scale domain-specific KBs (e.g., \cite{ernst-bmc15,hope-naacl21}). These KBs have a variety of uses: they lead to ease of information access by domain researchers \cite{tsatsaronis-bmc15,hamon-sw17}, provide data for developing domain-specific ML models \cite{nadkarni-akbc21}, and potentially help in accelerating scientific discoveries \cite{materials_project, challenges_21}. 

Significant research exists on IE from \emph{text} of research papers (see \citet{nasar-scientometrics18} for a survey), but less attention is given to IE (often, numeric) from \emph{tables}. Tables may report the performance of algorithms on a dataset, quantitative results of clinical trials, or other important information. Of special interest to us are tables that mention the composition and properties of an entity. Such tables are ubiquitous in various fields such as food and nutrition (tables of food items with nutritional values, see Tables 1-4 in \citet{Cavalcanti2021DonkeyMA} and Table 2 in \citet{Stokvis2021EvaluationOT}), fuels (constituents and calorific values, see Table 2 in \citet{Kar2022EffectOF} and \citet{Beliavskii2022EffectOF}), building construction (components and costs, see Table 4 in \citet{Aggarwal2022ComponentRC}), materials (constituents and properties, see Table 1 and 2 in \citet{kasimuthumaniyan2020understanding} and  Table 4 in \citet{keshri2022elucidating}), medicine (compounds with weights in drugs, see Table 1 in \citet{kalegari2014chemical}), and more. 

\begin{figure*}[t]
\begin{center}
    \centering
    \includegraphics[width=\textwidth]{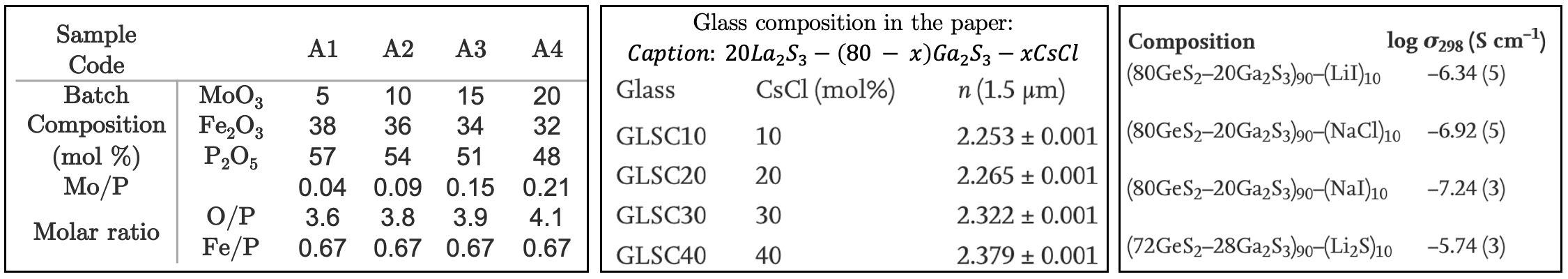}
    \caption{Examples of composition tables (a) Multi-cell complete-info \cite{fig_1a_mcc_all} (b) Multi-cell partial-info {with caption on top} \cite{fig_1b_mcc_pi} (c) Single-cell \cite{fig_1c_scc}}
    \label{fig:three comp_tables}
    \vspace{-0.6cm}
\end{center}
\end{figure*}

In materials science (MatSci) articles, the details on synthesis and characterization are reported in the text~\cite{mysore-etal-2019-materials}, while material compositions are mostly reported in tables~\cite{jensen2019machine}.
A preliminary analysis of MatSci papers reveals that $\sim$85\%\footnote{estimated by randomly choosing 100 compositions from a MatSci database and checking where they are reported} of material compositions and their associated properties (e.g., density, stiffness) are reported in tables and not text. 
Thus, IE from tables is essential for a comprehensive understanding of a given paper, and for increasing the coverage of resulting KBs. To this extent, we define a novel NLP task of extraction of materials (via IDs mentioned in the paper), constituents, and their relative percentages. For instance, Fig.\ref{fig:three comp_tables}a should output four materials A1-A4, where ID A1 is associated with three constituents (MoO$_3$, Fe$_2$O$_3$, and P$_2$O$_5$) and their respective percentages, 5, 38, and 57. A model for this task necessitates solving several challenges, which are discussed in detail in Sec.~\ref{sec:problem}. While many of these issues have been investigated separately, e.g., numerical IE \cite{madaan-aaai16}, unit extraction \cite{sarawagi14}, chemical compound identification \cite{weston2019named}, NLP for tables \cite{jensen2019machine,swain2016chemdataextractor}, solving all these in concert creates a challenging testbed for the NLP community. 


Here, we harvest a distantly supervised training dataset of 4,408 tables and 38,799 composition-constituent tuples by aligning a MatSci database with tables in papers. We also label 1,475 tables manually for dev and test sets. We build a baseline system \sys, which uses a pipeline of a domain-specific language model \cite{gupta_matscibert_2022}, and two graph neural networks (GNNs), along with several hand-coded features and constraints. We evaluate our system on accuracy metrics for various subtasks, including material ID prediction, tuple-level predictions, and material-level complete predictions. We find that \sys's GNN architecture obtains a 7-15 points increase in accuracy numbers, compared to table processors \cite{herzig-etal-2020-tapas, yin-etal-2020-tabert}, which linearize the table for IE. 
Subsequent analysis reveals common sources of \sys\ errors, which will inform future research. We release all our data and code\footnote{\url{https://github.com/M3RG-IITD/DiSCoMaT}} for further research on this challenging task.

\section{Related work} 
\label{sec:related_work}
\vspace{-0.075in}
Recent works have developed neural models for various NLP tasks based on tabular data, \textit{viz}, tabular natural language inference \cite{tnli_orihuela2021natural, xinfotabs}, QA over one or a corpus of tables \cite{herzig-etal-2020-tapas, yin-etal-2020-tabert, tabnet_Arik_Pfister_2021, feifei_row_col_interact, pan-etal-2021-cltr, chemmengath-emnlp21}, table orientation classification \cite{deeptable_habibi2020,nishida2017understanding}, and relation extraction from tables \cite{relation_ex_tables_2013, neurel_relex_2}. Several recent papers study QA models---they all linearize a table and pass it to a pre-trained language model. For example, 
\textsc{TaPas} \cite{herzig-etal-2020-tapas} does this for Wikipedia tables to answer natural language questions by selecting table cells and aggregation operators. \textsc{TaBERT} \cite{yin-etal-2020-tabert} and RCI \cite{feifei_row_col_interact} also use similar ideas alongside some architectural modifications to handle rows and columns better. \textsc{TABBIE} \cite{tabbie} consists of two transformers that encode rows and columns independently, whereas \textsc{TaPEx} uses encoder-decoder architecture using BART. TABBIE and \textsc{TaPEx} also introduce pre-training over tables to learn table representations better. Similar to our work, tables have also been modeled as graphs for sequential question answering over tables \cite{muller_acq_2019}. However, all these works generally assume a fixed and known structure of tables with the same orientation, with the top row being the header row in all cases -- an assumption violated in our setting. 
\\
{\bf Orientation and semantic structure classification:} 
DeepTable \cite{deeptable_habibi2020} is a permutation-invariant neural model, which classifies tables into three orientations, while TabNet \cite{nishida2017understanding} uses RNNs and CNNs in a hybrid fashion to classify web tables into five different types of orientations. \textsc{InfoTabS} \cite{infotabs} studies natural language inference on tabular data via linearization and language models, which has been extended to the multilingual setting \cite{xinfotabs}, and has been combined with knowledge graphs \cite{transkblstm}. Some earlier works also focused on annotating column types, entity ID cells, and pair of columns with binary relations, based on rule-based and other ML approaches, given a catalog~\cite{sunita_iitb}.

\section{Challenges in composition extraction from tables} 
\label{sec:problem}



We analyze numerous composition tables in  MatSci research papers (see Figures \ref{fig:three comp_tables}, \ref{fig:regex_var_comp_table} and \ref{fig:two_comp_tables_2} for examples), and find that the task has several facets, with many table styles for similar compositions. We now describe the key challenges involved in the task of composition extraction from tables.\\
     $\bullet$ \textbf{Distractor rows and columns}: Additional information such as material properties, molar ratios, and std errors in the same table. E.g., in Figure \ref{fig:three comp_tables}a, the last three rows are distractor rows.\\ 
    $\bullet$ \textbf{Orientation of tables}: Table shown in Figure \ref{fig:two_comp_tables_2}a is a row oriented table---different compositions are written in different rows. The table in Figure \ref{fig:three comp_tables}a is a column-oriented table.\\
    $\bullet$ \textbf{Different units}: Compositions can be in different units such as mol\%, weight\%,  mol fraction, weight fraction. Some tables express composition in both molar and mass units.\\
    $\bullet$ \textbf{Material IDs}: Authors refer to different materials in their publication by assigning them unique IDs. These material IDs may not be specified every time, (e.g., Fig.~\ref{fig:three comp_tables}c).\\ 
    $\bullet$ \textbf{Single-cell compositions (SCC)}: In Fig.~\ref{fig:three comp_tables}a, all compositions are present in multiple table cells. Some authors report the entire composition in a single table cell, as shown in Fig. \ref{fig:three comp_tables}c.\\
    $\bullet$ \textbf{Percentages exceeding 100:} Sum of coefficients may exceed 100, and re-normalization is needed. A common case is when a dopant is used; its amount is reported in excess.\\
    $\bullet$ \textbf{Percentages as variables}: Contributions of constituents may be expressed using variables like $x, y$. In Fig. \ref{fig:regex_var_comp_table} (see App. A), x represents the mol\% of (GeBr$_4$) and the 2$^{\rm{nd}}$ row contains its value.\\
    $\bullet$ \textbf{Partial-information tables}: It is also common to have percentages of only some constituents in the table; the remaining composition is to be inferred based on paper text or table caption,  
    e.g., Figure \ref{fig:three comp_tables}b. Another example: if the paper is on silicate glasses, then SiO$_2$ is assumed.\\   
    $\bullet$ \textbf{Other corner cases}: There are several other corner cases like percentages missing from the table, compounds with variables (e.g., R$_2$O in the header; the value of R to be inferred from material ID), and highly unusual placement of information (some examples in appendix).\\
\section{Problem formulation}
\label{problem}
Our goal is automated extraction of material compositions from tables.  Formally, given a table $T$, its caption, and the complete text of publication in which $T$ occurs, we aim to extract compositions expressed in $T$, in the form $\{(id, c_k^{id}, p_k^{id}, u_k^{id} )\}_{k=1}^{K^{id}}$. Here, $id$ represents the material ID, as used in the paper. Material IDs are defined by MatSci researchers to succinctly refer to that composition in text and other tables. $c_k^{id}$ is a constituent element or compound present in the material, $K^{id}$ is the total number of constituents in the material, $p_k^{id} > 0$ denotes the percentage contribution of $c_k^{id}$ in its composition, and $u_k^{id}$ is the unit of $p_k^{id}$ (either mole\% or weight\%). For instance, the desired output tuples corresponding to ID A1 from Figure \ref{fig:three comp_tables}a are (A1, MoO$_3$, 5, mol\%), (A1, Fe$_2$O$_3$, 38, mol\%), (A1, P$_2$O$_5$, 57, mol\%).

\section{Dataset construction}
\label{sec:dataset_construction}
We match a MatSci DB of materials and compositions with tables from published papers, to automatically provide distantly-supervised labels for extraction. 
We first use a commercial DB \cite{interglad} of glass compositions with the respective references. Then, we extract all tables from the 2,536 references in the DB using text-mining API \cite{elsevier_api}. We use a table parser \cite{mit_table_parser} for raw XML tables and captions. This results in 5,883 tables of which 2,355 express compositions with 16,729 materials, and 58,481 (material ID, constituent, composition percentage, unit) tuples.  We keep tables from 1,880 papers for training, and the rest are split into dev and test (see Table \ref{tab:dataset_stats_2}).

The DB does not contain information about the location of a given composition in the paper -- in text, images, graphs, or tables. If present in a table, it can appear in any column or row. Since we do not know the exact location of a composition, we use distantly supervised train set construction \cite{mintz-etal-2009-distant}. First, we simply match the chemical compounds and percentages (or equivalent fractions) mentioned in the DB with the text in a table from the associated paper. If all composition percentages are found in multiple cells of the table, it is marked as MCC-CI (multi-cell composition with complete information). However, due to several problems (see Appendix~\ref{sec:problem}), it misses many composition tables. To increase the coverage, we additionally use a rule-based composition parser (described below), but restricted to only those compounds (CPD non-terminal in Figure \ref{fig:comp_parser}) that appear in the DB for this paper. 

Our distant supervision approach obtains table-level annotation (NC, SCC, MCC-PI, MCC-CI), where a table is labeled as non-composition, single/multi cell composition with partial/complete information. It also obtains annotation for each row or column into four labels: ID, composition, constituent, and other. While training data is created using distant supervision, dev and test sets are hand annotated.
We now explain the dataset construction process in further detail. 


{\textbf{Rule-based composition parser: }
The parser helps find names of constituents from MCC tables, and also match full compositions mentioned in SCC tables. 
Recall that in SCC tables, the full \emph{composition expression} is written in a single cell in the row/column corresponding to each Material ID. Such compositions are close to regular languages and can be parsed via regular expressions. 

\begin{wrapfigure}{l}{0.3\textwidth}
    \vspace{-0.1in}
    \centering
    \FramedBox{2.5cm}{5.3cm}
    {\small{CMP = PAT$_1$ | PAT$_2$ | PAT$_3$\\
    PAT$_i$ = START CST$_i$ (SEP CST$_i$)$^+$ END \\
    ~\\
    CST$_1$ = NUM? W CPD\\
    CST$_t$ = CST$_1$ (SEP CST$_1$)$^*$ \\
    CST$_2$ = (CST$_t$ | OB CST$_t$ CB) W NUM  \\
    CST$_3$ = NUM W (CST$_t$ | OB CST$_t$ CB)
    }}
    \caption{Regexes in parser}
        \label{fig:comp_parser}
        \vspace{-2ex}
\end{wrapfigure}

Figure \ref{fig:comp_parser} shows the regular expression (simplified, for understandability) used by the parser. Here CMP denotes the matched composition, PATs are the three main patterns for it, CSTs are sub-patterns, CPD is a compound, NUM is a number, and OB and CB are, respectively, open and closed parentheses (or square brackets). W is zero or more whitespace characters, and SEP contains explicit separators like `-' or `+'. START and END are indicators to separate a regular expression from the rest of the text.

The first pattern parses simple number-compound expressions like 40Bi$_2$O$_3$ * 60B$_2$O$_3$. Here each of the two constituents will match with CST$_1$. The other two patterns handle \emph{nested} compositions, where simple expressions are mixed in a given ratio. The main difference between the second and third patterns is in the placement of outer ratios -- after or before the simple composition, respectively. Example match for PAT$_2$ is (40Bi$_2$O$_3$+60B$_2$O$_3$)30 - (AgI+AgCl)70, and for PAT$_3$ is 40Bi$_2$O$_3$,40B$_2$O$_3$,20(AgI:2AgCl).

To materialize the rules of the rule-based composition parser, we pre-label compounds. For our dataset, we use a list-based extractor, though other chemical data extractors \cite{chemdataextractor_v1} may also be used.  After parsing, all coefficients are normalized so that they sum to hundred. For nested expressions, the outer ratio and the inner ones are normalized separately and then multiplied.

The compositions parsed by rule-based composition parser are then matched with entries in the DB. A successful matching leads to a high-quality annotation of composition expressions in these papers. If this matching happens: (i) in a single cell, the table is deemed as SCC, (ii) on caption/paper text that has an algebraic variable (or compound) found in the table, it is marked as MCC-PI (see Figure \ref{fig:three comp_tables}(b)). In case of no matching, the table is marked as NC. This automatic annotation is post-processed into row, column and edge labels.

One further challenge is that material IDs mentioned in papers are not provided in the DB. So, we manually annotate material IDs for all the identified composition tables in the training set. This leads to a train set of 11,207 materials with 38,799 tuples from 4,408 tables. Since the train set is distantly supervised and can be noisy, two authors (one of them is a MatSci expert) of this paper manually annotated the dev and test tables with row/column/edge labels, units, tuples, compositions, and table type, resulting in over 2,500 materials and over 9,500 tuples per set. We used Cohen's Kappa measure for identifying inter-annotator agreement, which was 86.76\% for Glass ID, 98.47\% for row and column labels, and 94.34\% for table types. Conflicts were resolved through mutual discussions. Further statistics and the description of the developed in-house annotation tools used for manual annotations are discussed in ~\ref{app:dataset-details}.

\section{\sys\ architecture} 
\label{sec:model}

\begin{figure}[hbt]
    \centering
    \includegraphics[width=0.45\textwidth, height=3cm]{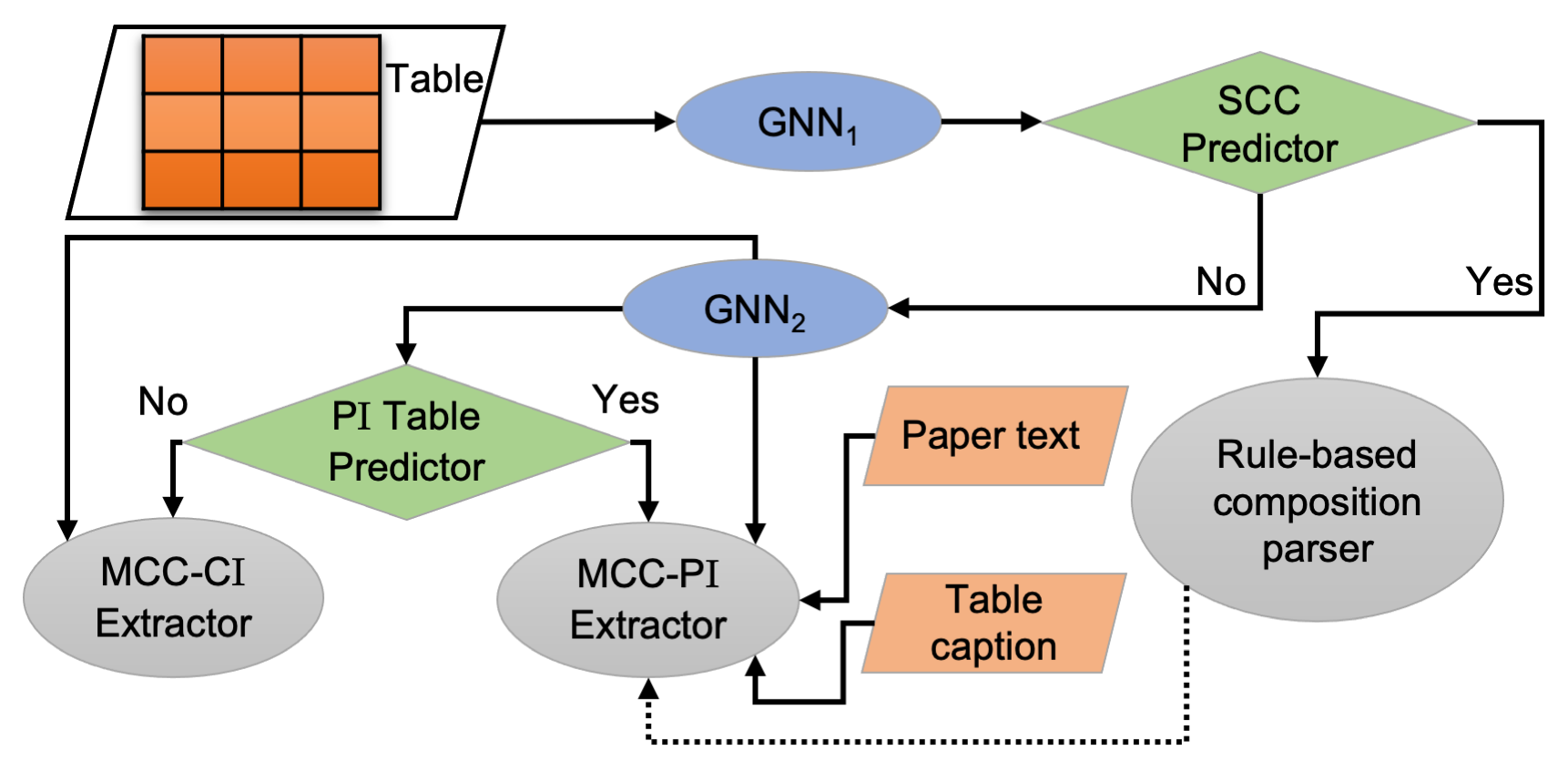}
    \caption{The design of \sys}
    \label{fig:comp_table_cls}
    \vspace{-1ex}
\end{figure}

Figure \ref{fig:comp_table_cls} illustrates the basic pipeline for extraction in \sys. We find that the simplest task is to identify whether the table $T$ is an SCC table, owing to the distinctive presence of multiple numbers, and compounds in single cells. \sys{} first runs a GNN-based SCC predictor, which classifies $T$ as an SCC table or not. For the SCC table, it uses the rule-based composition parser (described in Sec. \ref{sec:dataset_construction}). For the other category, \sys{} runs a second GNN (GNN$_2$), and labels rows and columns of $T$ as compositions, material IDs, constituents, and others. If no constituents or composition predictions are found, then $T$ is deemed to be a non-composition (NC) table. Else, it is an MCC table, for which \sys{} predicts whether it has all information in $T$ or some information is missing (partial-information predictor). If it is a complete information table, then GNN$_2$'s predictions are post-processed into compositions. If not, the caption and text of the paper are also processed, along with GNN$_2$'s predictions leading to final composition extraction. We note that our system ignores statistically infrequent corner cases, such as single-cell partial information tables -- we discuss this further in our error analysis. We now describe each of these components, one by one.

\subsection{GNN$_1$ and GNN$_2$ for table processing}
\label{sec:gnn}
At the core of \sys\ are two GNNs that learn representations for each cell, row, column and the whole table. Let table $T$ has $R$ rows, $C$ columns, and text at $(i,j)^{th}$ cell be $t_{ij}$, where $1\leq i\leq R$, and $1\leq j\leq C$. We construct a directed graph 
$G_T = (V_T, E_T)$, where $V_T$ has a node for each cell $(i,j)$, one additional node for each row and column, denoted by $(i,0)$ and $(0,j)$, respectively, and one node for the whole table represented by $(0,0)$. There are bidirectional edges between two nodes of the same row or column. All cell nodes have directed edges to the table node and also their corresponding row and column nodes. 
The table, row, and column embeddings are randomly initialized with a common vector, which gets trained during learning. A node $(i,j)$'s embedding $\overrightarrow{x_{ij}}$ is initialized by running a language model $LM$ over $t_{ij}$.

As constructed, $G_T$ is permutation-invariant, i.e., if we permute rows or columns, we get the same graph and embeddings. However, initial rows/columns can be semantically different, since they often represent headings for the subsequent list. For instance, material IDs are generally mentioned in the first one or two rows/columns of the table. So, we additionally define \emph{index embeddings} $\overrightarrow{p_{i}}$ to represent a row/column numbered $i$. We use the same index embeddings for rows and columns so that our model stays transpose-invariant. We also observe that while first few indices are different, the semantics is generally uniform for indices higher than 3. 
Accordingly, to allow \sys\ to handle large tables, we simply use $\overrightarrow{p_{i}}=\overrightarrow{p_{3}}~ \forall i > 3$. 
Finally, any manually-defined features added to each node are embedded as $\overrightarrow{f}$ and concatenated to the cell embeddings. Combining all ideas, a cell embedding is initialized as: 
\vspace{-1ex}
\[
\overrightarrow{x_{ij}} =  \overrightarrow{f_{ij}} \;||\; \left( LM_{CLS}(\langle CLS ~ t_{ij} ~ SEP \rangle) \!+\! \overrightarrow{p_{i}} \!+\! \overrightarrow{p_{j}} \right) \] 
Here, $1\leq i \leq R, 1 \leq j \leq C$. $||$ is the concat operation and $LM_{CLS}$ gives the contextual embedding of the $CLS$ token after running a $LM$ over the sentence inside $\langle\rangle$. Message passing is run on the graph $G_T$ using a GNN, which computes a learned feature vector $\overrightarrow{h}$ for every node: \[\{\overrightarrow{h_{ij}}\}_{i,j=(0,0)}^{(R,C)} = GNN\left(\{\overrightarrow{x_{ij}}\}_{i,j=(0,0)}^{(R,C)}\right).\]

\subsection{SCC Predictor}
\label{sec:single_cell_table}
In its pipeline, \sys\ first classifies whether $T$ is an SCC table. 
For that, it runs a GNN (named GNN$_1$) on $T$ with two manually defined features (see below). 
It then implements a Multi-layer Perceptron MLP$_1$ over the table-level feature vector $\overrightarrow{h_{00}}$  to make the prediction. Additionally, GNN$_1$ also feeds row and column vectors $\overrightarrow{h_{i0}}$ and $\overrightarrow{h_{0j}}$ through another MLP (MLP$_2$) to predict whether they contain material IDs  or not. If $T$ is predicted as an SCC table, then one with the highest MLP$_2$ probability is deemed as material ID row/column (provided probability > $\alpha$, where $\alpha$ is a hyper-parameter tuned on dev set), and its contents are extracted as potential material IDs. If all row and column probabilities are less than $\alpha$, then the table is predicted to not have Material IDs, as in Figure \ref{fig:three comp_tables}c.


For an SCC table, \sys\ must parse the full \emph{composition expression} written in a single cell in the row/column corresponding to each Material ID, for which it makes use of the rule-based composition parser (as described in Section \ref{sec:dataset_construction}). The only difference is that at test time there is no DB available and hence extracted compositions cannot be matched with further. Consequently, \sys\ retains all extracted composition expresssions from the parser for further processing. 

For units, \sys\ searches for common unit keywords such as mol, mass, weight, and their abbreviations like wt.\%, and at.\%. The search is done iteratively with increasing distance from the cell containing the composition. If not found in the table, then the caption is searched. If still not found, mole\% is used as default.

\textbf{Manual Features:} GNN$_1$ uses two hand-coded features. The first feature is set to true if that cell contains a composition that matches our rule-based composition parser. Each value, true or false, is embedded as $\overrightarrow{o}$.
The second feature named \emph{max frequency feature} adds the bias that material IDs are generally unique in a table. We compute $q_{i}^{r}$ and $q_{j}^{c}$, which denote the maximum frequency of any non-empty string occurring in the cells of row $i$ and column $j$, respectively. If these numbers are on the lower side, then that row/column has more unique strings, which should increase the probability that it contains material IDs. The computed $q$ values are embedded in a vector as $\overrightarrow{q}$. The embedded feature $\overrightarrow{f_{ij}}$ for cell $(i,j)$ is initialized as $\overrightarrow{o_{ij}} \;||\;  (\overrightarrow{q_{ij}^r}+\overrightarrow{q_{ij}^c})$.

\subsection{MCC-CI and MCC-PI Extractors}
\label{sec:multi-cell}

If $T$ is predicted to not be an SCC table, \sys\ runs it through another GNN (GNN$_2$). The graph structure is very similar to $G_T$ from Section \ref{sec:gnn}, but with two major changes. 
First, a new \emph{caption node} is created with initial embedding as given by $LM$ processing the caption text.  
Edges are added from the caption node to all row and column nodes. To propagate the information further to cells, edges are added from row/column nodes to corresponding cell nodes. The caption node especially helps in identifying non-composition (NC) tables. Second,
the max frequency feature from Section \ref{sec:single_cell_table} is also included in this GNN. 

We use tables in Figure \ref{fig:two_comp_tables_2} as our running examples. While Figure \ref{fig:two_comp_tables_2}a is a complete-information table, Figure \ref{fig:two_comp_tables_2}b is not, and can only be understood in the context of its caption, which describes the composition as  $[(Na_2O)_x(Rb_2O)_{1-x}]_y(B_2O_3)_{1-y}$. Here ${x}$ and ${y}$ are variables, which also need to be extracted and matched with the caption. \sys\ first decodes the row and column feature vectors $\overrightarrow{h_{i0}}$ and $\overrightarrow{h_{0j}}$, as computed by GNN$_2$, via an MLP$_3$ into four classes: composition, constituent, ID, and other (label IDs 1, 2, 3, 0, respectively). The figures illustrate this labelling for our running example. The cell at the intersection of composition row/column and constituent column/row represents the percentage contribution of that constituent in that composition. 

\begin{figure}[t]
\begin{center}
    \centering
    \begin{subfigure}[b]{0.4\textwidth}
        \centering
        \includegraphics[width=\textwidth, height=2.8cm]{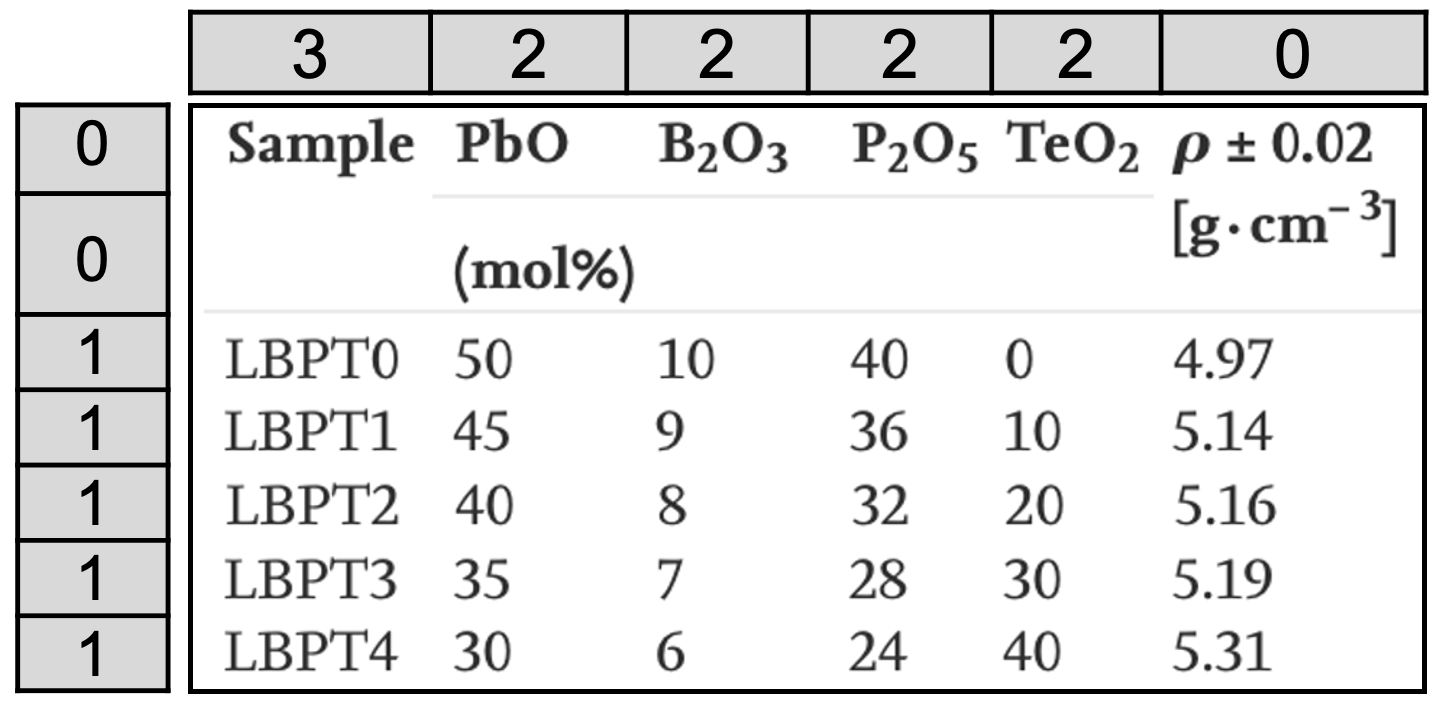}
    \end{subfigure}
    \hfill
    \begin{subfigure}[b]{0.4\textwidth}
        \centering
        \includegraphics[width=\textwidth, height=2.8cm]{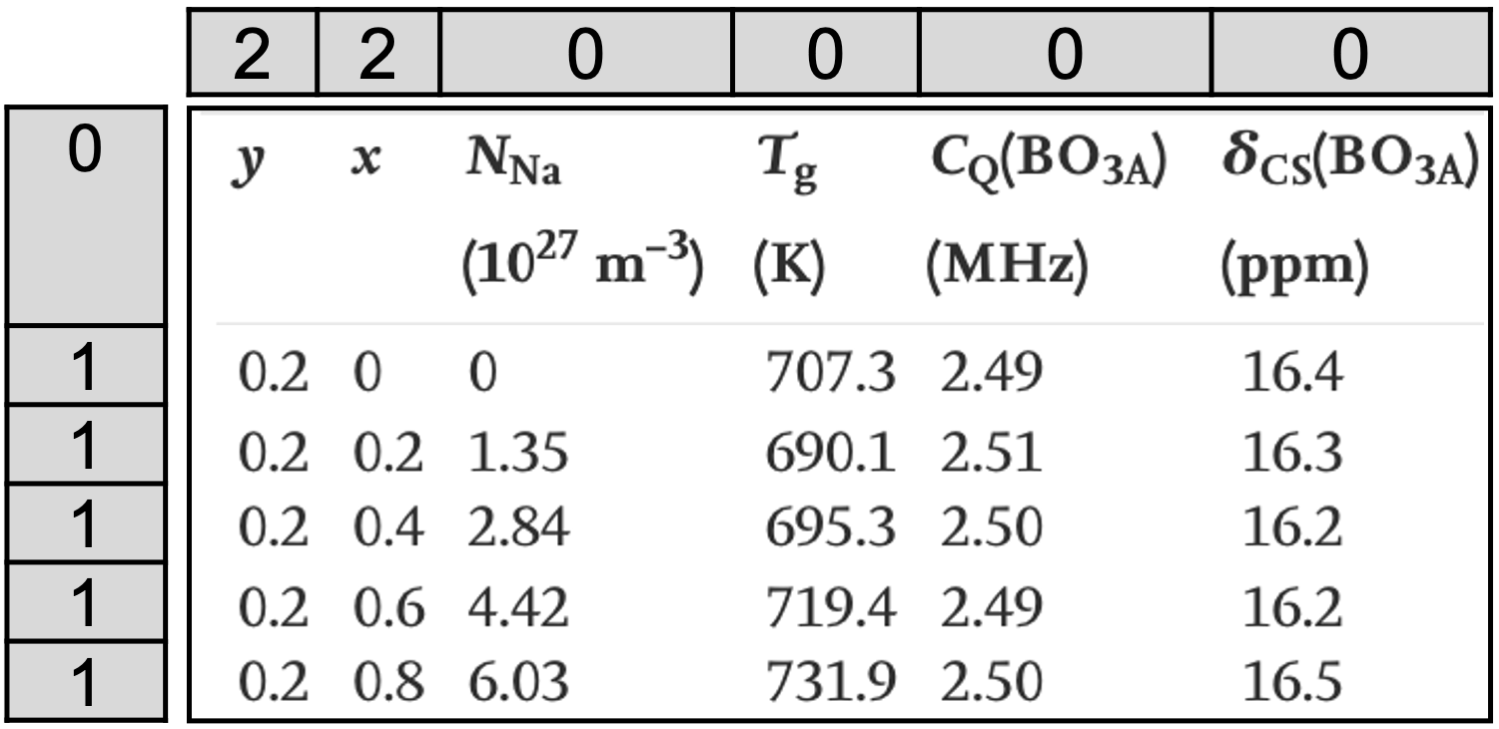}
    \end{subfigure}
    \caption{Multi-cell composition tables (a) Complete information \cite{fig_5a} (b) Partial information \cite{fig_5b}}
    \label{fig:two_comp_tables_2}
\end{center}
    \vspace{-4ex}
\end{figure}

Further, to associate the identified percentage contribution with the corresponding constituent (like P$_2$O$_5$ in Figure \ref{fig:two_comp_tables_2}a) or variables ${x}$ and ${y}$ in Figure \ref{fig:two_comp_tables_2}b), we perform classification at the edge level. For ease of exposition, we describe our method in this Section \ref{sec:multi-cell} for the setting that the table has been predicted by GNN$_2$ to have row-wise orientation, i.e., rows are compositions and columns are constituents. A transposed computation is done in the reverse case. Since the constituent/variable will likely occur in the same column or row as the cell containing percentage contribution, our method computes an edge feature vector: for edge $e=(i,j)\rightarrow(i',j')$, s.t. $i=i' \vee j=j' $, the feature vector $\overrightarrow{h_e}= \overrightarrow{h_{ij}} \;||\; \overrightarrow{h_{i'j'}}$.
It then takes all such edges $e$ from cell $(i,j)$, if row $i$ is labeled composition and column $j$ is labeled constituent. Each edge $e$ is classified through an MLP$_4$, and the edge with the maximum logit value is picked to identify the constituent/variable. This helps connect 36 to P$_2$O$_5$ and 0.8 to ${x}$ in our running examples. 
GNN$_2$ also helps in predicting NC tables. In case none of the rows/columns are predicted as 1 or 2, then the table is deemed as NC and discarded.

\textbf{Partial information table predictor:}
Next, \sys\ distinguishes between complete-information (CI) and partial-information (PI) MCC tables. It uses a logistic regression model with custom input features for this prediction task. 
Let $P$ and $Q$ be the sets of all row indices with label 1 (composition) and column indices with label 2 (constituent), respectively. Also, assume $n_{ij}$ is the number present in table cell $(i, j)$ or $0$ if no number is present.  To create the features, we first extract all the constituents (compounds) and variables predicted by MLP$_4$. We now construct five table-level features (F1-F5). F1 and F2 count the number of unique variables and chemical compounds extracted by MLP$_4$. The intuition is that if F1 is high, then it is more likely an MCC-PI, and vice-versa if F2 is high. F3 computes the number of rows and columns labeled as 2 (constituent) by MLP$_3$. The more the value of F3, the more likely it is that the table is MCC-CI. Features F4 (and F5) compute the maximum (average) of the sum of all extracted compositions. The intuition of F4 and F5 is that the higher these feature values, the higher the chance of the table being an MCC-CI. Formally, 

\[
\mathrm{F4}=(\max_{i \in P} \sum_{j \in Q}n_{ij}) ~~~~~~~~~~~~~
\mathrm{F5}=(\frac{1}{|P|} \sum_{i \in P} \sum_{j \in Q}n_{ij}).
\]

\textbf{MCC table extractor:} For MCC-CI, MLP$_3$ and MLP$_4$ outputs are post-processed, and units are added (similar to SCC tables), to construct final extracted tuples. For MCC-PI, on the other hand, information in the text needs to be combined with the MLP outputs for final extraction. The first step here is to search for the composition expression, which may be present in the table caption, table footer, and if not there, somewhere in the rest of the research paper. Here, \sys\ resorts to using our rule-based composition parser from Figure \ref{fig:comp_parser}, but with one key difference. Now, the composition may contain variables ($x$, $y$) and even mathematical expressions like $100-x$. So the regular grammar is enhanced to replace the non-terminal NUM with a non-terminal EXPR, which represents, numbers, variables, and simple mathematical expressions over them. An added constraint is that if there are variables in set $Q$, then those variables must be present in the matched composition expression. \sys\ completes the composition by substituting the variable values from every composition row into the matched composition. There may be other types of MCC-PI tables where only compounds are identified in tables, such as Figure \ref{fig:three comp_tables}b. For these, \sys\ first computes the constituent contributions in terms of variables from the composition expression, and then equates it with the numbers present in rows/columns labeled 1 (composition). In our example, \sys\ matches ${x}$ with the numbers 10, 20, 30, and 40, and the rest of the composition is extracted by processing the composition expression in the caption with these values of ${x}$. Units and material IDs are added to the tuples, similar to other tables.


\subsection{Constraint-aware loss functions}
\label{sec:constraints}

\sys\ needs to train the two GNNs and the PI table predictor. Our data construction provides gold labels for each prediction task (discussed in the next section), so we train them component-wise. The PI table predictor is trained on standard logistic regression loss. GNN$_1$ is trained on a weighted sum of binary cross entropy loss for SCC table classification and row/column classification for material IDs --  weight is a hyper-parameter. Similarly, the GNN$_2$ loss function consists of the sum of row/column cross-entropy and edge binary cross-entropy losses. 

GNN$_2$ has a more complex prediction problem since it has to perform four-way labeling for each row and column. In initial experiments, we find that the model sometimes makes structural errors like labeling one row as a constituent and another row as a composition in the same table -- highly unlikely as per the semantics of composition tables. To encourage GNN$_2$ to make structurally consistent predictions, we express a set of constraints on the complete labelings, as follows. (1) A row and a column cannot both have compositions or constituents.  
(2) Composition and material ID must be orthogonally predicted (i.e, if a row has a composition then ID must be predicted in some column, and vice versa). (3) Constituents and material IDs must never be orthogonally predicted (if rows have constituents then another row must have the ID). And, (4) material ID must occur at most once for the entire table. 
As an example, constraint (1) can be expressed as a hard constraint as:
\begin{center}
\(
r_i = l \Rightarrow c_j \neq l ~~~~\forall i\in\{1,\ldots, R\}, j\in \{1,\ldots,C\}, l\in\{1,2\}. 
\)
\end{center}
Here, $r_i$ and $c_j$ are predicted labels of row $i$ and column $j$. We wish to impose these structural constraints at training time so that the model is trained to honor them. We follow prior work by Nandwani et al. \cite{yatin}, to first convert these hard constraints into a probabilistic statement. For example, constraint (1) gets expressed as:
\begin{center}
\(
P(r_i = l; \theta) + P(c_j = l; \theta) - 1 \leq 0\) \( \forall i \in \{1, \dots, R\}, \; j \in \{1, \dots, C\}, \; l \in \{1, 2\}.
\)
\end{center}
$\theta$ represents GNN$_2$'s parameters. Following the same work, each such constraint gets converted to an auxiliary penalty term, which gets added to the loss function for constraint-aware training. The first constraint gets converted to: 
\[ \lambda \sum_{i=1}^R \sum_{j=1}^C \sum_{l=1}^2 max(0, P(r_i = l; \theta) + P(c_j = l; \theta) - 1) \]\\
This and similar auxiliary losses for other constraints (App.~\ref{app:constraint}) get added to the GNN$_2$'s loss function for better training. $\lambda$ is a hyper-parameter. We also use constraint (4) for GNN$_1$ training.

\section{Experiments} 
\label{sec:experiments}

\textbf{Baseline models: }
We implement \sys\ with $LM$ as \textsc{MatSciBERT} \cite{gupta_matscibert_2022}, and the GNNs as Graph Attention Networks \cite{velickovic2018graph}. We compare \sys\ with six non-GNN baseline models. Our first baseline is \textsc{TaPas} \cite{herzig-etal-2020-tapas}, a state-of-the-art table QA system, which flattens the table, adds row and column index embeddings, and passes as input to a language model. To use TaPas for our task, we use table caption as a proxy for the input question. All the model parameters in this setting are initialized randomly. 
Next, we use \textsc{TaBERT} \cite{yin-etal-2020-tabert}, which is a pretrained LM that jointly learns representations for natural (NL) sentences and tables by using pretraining objectives of masked column prediction (MCP) and cell value recovery (CVR). It finds table cell embeddings by passing row linearizations concatenated with the NL sentence into a language model and then applying vertical attention across columns for information propagation.
Finally, we use \textsc{TABBIE}, which is pretrained by corrupt cell detection and learns exclusively from tabular data without any associated text, unlike the previous baselines. 
Additionally, we replace the LM of all models with \textsc{MatSciBERT} to provide domain-specific embeddings to obtain the respective \textsc{Adapted} versions. \\
We also implement a simple rule-based baseline for MCC-CI and NC tables. The baseline identifies
constituent names using regex matching and a pre-defined list of compounds, extracts numbers from cells and finds the units using simple heuristics to generate the required tuples. Further details on baselines is provided in App.~\ref{app:baseline}.

\begin{table*}[htb]
\centering
\scalebox{0.85}{
\begin{tabular}{lcccc}
    \toprule
    \textbf{Model} & \textbf{ID F$_1$} & \textbf{TL F$_1$} & \textbf{MatL F$_1$} &  \textbf{CV} \\
    \midrule
    
    \textsc{TaPas} & 80.37 ($\pm$ 4.78) & 71.23 ($\pm$ 0.77) & 49.88 ($\pm$ 0.10) & 543.67 \\
    \textsc{TaPas-Adapted} & \textbf{89.65} ($\pm$ 0.46) & 70.91 ($\pm$ 3.79) & 57.88 ($\pm$ 2.73) & 490.33 \\
    \textsc{TaBERT} & 79.61 ($\pm$ 8.25) & 58.20 ($\pm$ 1.79) & 47.05 ($\pm$ 1.50) & 1729.67 \\
    \textsc{TaBERT-Adapted} & 85.07 ($\pm$ 6.28) & 59.31 ($\pm$ 0.67) & 50.10 ($\pm$ 2.86) & 1195.67 \\
    \textsc{TABBIE} & 80.99 ($\pm$ 2.41) & 50.90 ($\pm$ 3.34) & 47.03 ($\pm$ 2.14) & \textbf{388.00} \\
    \textsc{TABBIE-Adapted} & 80.18 ($\pm$ 5.38) & 53.20 ($\pm$ 5.57) & 48.89 ($\pm$ 2.73) & 728.67 \\
    \textsc{Rule Based System} & 72.64 & 54.44 & 47.38 & 0 \\
    \textsc{v-}\sys & 77.38 ($\pm$ 12.21) & \textbf{76.52} ($\pm$ 2.37) & \textbf{64.71} ($\pm$ 3.45) & 626.33 \\
    \bottomrule
\end{tabular}
}
\caption{Performance of \textsc{v-}\sys{} vs baseline models on the subset of data containing only MCC-CI and NC table types.}
\label{tab:baselines}
\vspace{-0.10cm}
\end{table*}

    

    
\begin{table*}[htb]
\centering
\scalebox{0.85}{
\begin{tabular}{lccccc}
    \toprule
    \textbf{Model} & \textbf{TT Acc.} & \textbf{ID F$_1$} & \textbf{TL F$_1$} & \textbf{MatL F$_1$} & \textbf{CV} \\
    \midrule
    
    \sys & 88.35 ($\pm$ 1.20) & 84.57 ($\pm$ 2.16) & \textbf{70.04} ($\pm$ 0.69) & \textbf{63.53} ($\pm$ 1.45) & 75.22 \\
    \midrule

    \sys{} w/o features & \textbf{88.84} ($\pm$ 1.00) & 84.15 ($\pm$ 1.61) & 68.31 ($\pm$ 1.45) & 62.47 ($\pm$ 1.98) & 83.11 \\
    \sys{} w/o constraints & 88.47 ($\pm$ 0.31) & 84.07 ($\pm$ 0.83) & 69.68 ($\pm$ 1.21) & 61.44 ($\pm$ 1.00) & 434.44 \\
    \sys{} w/o captions & 87.35 ($\pm$ 0.71) & \textbf{84.76} ($\pm$ 0.68) & 66.82 ($\pm$ 1.90) & 62.68 ($\pm$ 3.33) & \textbf{17.89} \\
    \textsc{v-}\sys & 88.59 ($\pm$ 0.33) & 76.61 ($\pm$ 6.16) & 66.15 ($\pm$ 2.00) & 59.52 ($\pm$ 3.33) & 380.11 \\
    
    \bottomrule
\end{tabular}
}
\caption{Contribution of task-specific features and constraints in \sys\ on the complete dataset.}
\label{tab:ablations}
\vspace{-0.1in}
\end{table*}

\begin{table}[!htb]
\centering
\scalebox{0.75}{
\begin{tabular}{ccccc}
    \toprule
    \textbf{Table Type} & \textbf{ID F$_1$} & \textbf{TL F$_1$} & \textbf{MatL F$_1$} \\
    \midrule
    
    SCC & 88.81 ($\pm$ 1.54) & 79.89 ($\pm$ 0.18) & 78.21 ($\pm$ 0.14) \\
    
    MCC-CI & 93.91 ($\pm$ 1.46) & 77.62 ($\pm$ 1.07) & 65.41 ($\pm$ 4.35) \\
    
    MCC-PI & 70.67 ($\pm$ 11.58) & 50.60 ($\pm$ 2.59) & 51.66 ($\pm$ 2.21) \\

    \bottomrule
\end{tabular}
}
\caption{\sys{} performance on the table-types.}
\label{tab:perf_table_types}
\vspace{-0.25in}
\end{table}

\textbf{Evaluation metrics:} 
We compute several metrics in our evaluation. (1) \emph{Table-type (TT) prediction accuracy} -- it computes table-level accuracy on the 4-way table classification as NC, SCC, MCC-CI and MCC-PI. 
(2) \textit{ID F$_1$ score} computes F$_1$ score for Material ID extraction.  
(3) \textit{Tuple-level (TL) F$_1$ score} evaluates performance on the extraction of composition tuples. A gold is considered matching with a predicted 4-tuple if \emph{all} arguments match exactly. 
(4) \textit{Material-level (MatL) F$_1$ score} is the strongest metric. It evaluates whether all predicted information related to a material (including its ID, all constituents and their percentages) match exactly with the gold.
Finally, (5) \textit{constraint violations (CV)} counts the number of violations of hard constraints in the prediction. We consider all four types of constraints, as discussed in Section \ref{sec:constraints}. 
Implementation details are mentioned in App.~\ref{app:implementation}. 
\vspace{-1ex}
\subsection{Results}
\emph{How does table linearization compare with a graph-based model for our task?} To answer this question,  we compare \sys{} with four models that use linearization: {\sc TaPas}, \textsc{TaBERT}, and their adapted versions. \textsc{TaPas} and \textsc{TaBERT} do table level and row level linearizations respectively. Since the baselines do not have the benefit of regular expressions, features, and constraints, we implement a version of our model without these, which we call \textsc{v-}\sys. We do this comparison, trained and tested only on the subset of MCC-CI and NC tables since other table types require regular expressions for processing. As shown in Table \ref{tab:baselines} \textsc{v-}\sys{} obtain 6-7 pt higher F$_1$ on TL and MatL scores. Moreover, compared to the \textsc{Rule Based System}, \sys{} obtains upto 17 points improvement in the MatL F1 score.
This experiment suggests that a graph-based extractor is a better fit for our problem -- this led to us choosing a GNN-based approach for \sys. 


\emph{How does \sys{} perform on the complete task?}
Table \ref{tab:ablations}, reports \sys{} performance on the full test set with all table types. 
Its ID and tuple F$_1$-scores are 82 and 70, respectively. Since these errors get multiplied, unsurprisingly, its material-level F$_1$-score is lower (63.5). Table \ref{tab:perf_table_types} reports \sys{} performance for different table types. In this experiment, we assume that the table type is already known and run only the relevant part of \sys{} for extraction. We find that MCC-PI is the hardest table type since it requires combining information from text and tables for accurate extraction. A larger standard deviation in ID F$_1$ for MCC-PI is attributed to the fact that material IDs occur relatively rarely for this table type -- the test set for MCC-PI consists of merely 20 material ID rows and columns.

\emph{What is the incremental contribution of task-specific features and constraints?}
Table \ref{tab:ablations} also presents the ablation experiments. \sys{} scores much higher than \textsc{v-}\sys, which does not have these features and constraints. We also perform additional ablations removing one component at a time. Unsurprisingly constrained training helps with reducing constraint violations. Both constraints and features help with ID prediction, due to constraints (2), (3), (4) and max frequency feature. Removal of caption nodes significantly hurts performance on MCC-PI tables, as these tables require combining caption with table cells. Although the ablation study done by removing features, constraints, and captions individually does not show much of a difference on the tuple-level and material-level scores, we observe that on removing all the three factors, the performance of \textsc{v-}\sys{} drops significantly. Therefore, we can conclude that even though each component is improving the performance of \sys{} marginally, collectively, they help us to achieve significant gains.

\emph{What are the typical errors in \sys?} 
The confusion matrix in Figure \ref{fig:confusion} suggests that most table-type errors are between MCC-PI and NC tables. This could be attributed to the following reasons. (i) \sys{} has difficulty identifying rare compounds like Yb$_2$O$_3$, ErS$_{3/2}$, Co$_3$O$_4$ found in MCC-PI---these aren't present frequently in the training set. (ii) MCC-PI tables specify dopant percentages found in small quantities. 
(iii) Completion of composition in MCC-PI tables may require other tables from the same paper. (iv) Finally, MCC-PI composition may contain additional information such as properties that may bias the model to classify it as  NC. Some corner cases are given in App.~\ref{app:corner}. 
\begin{wrapfigure}{r}{0.25\textwidth}
    \centering
    \includegraphics[width=0.25\textwidth]{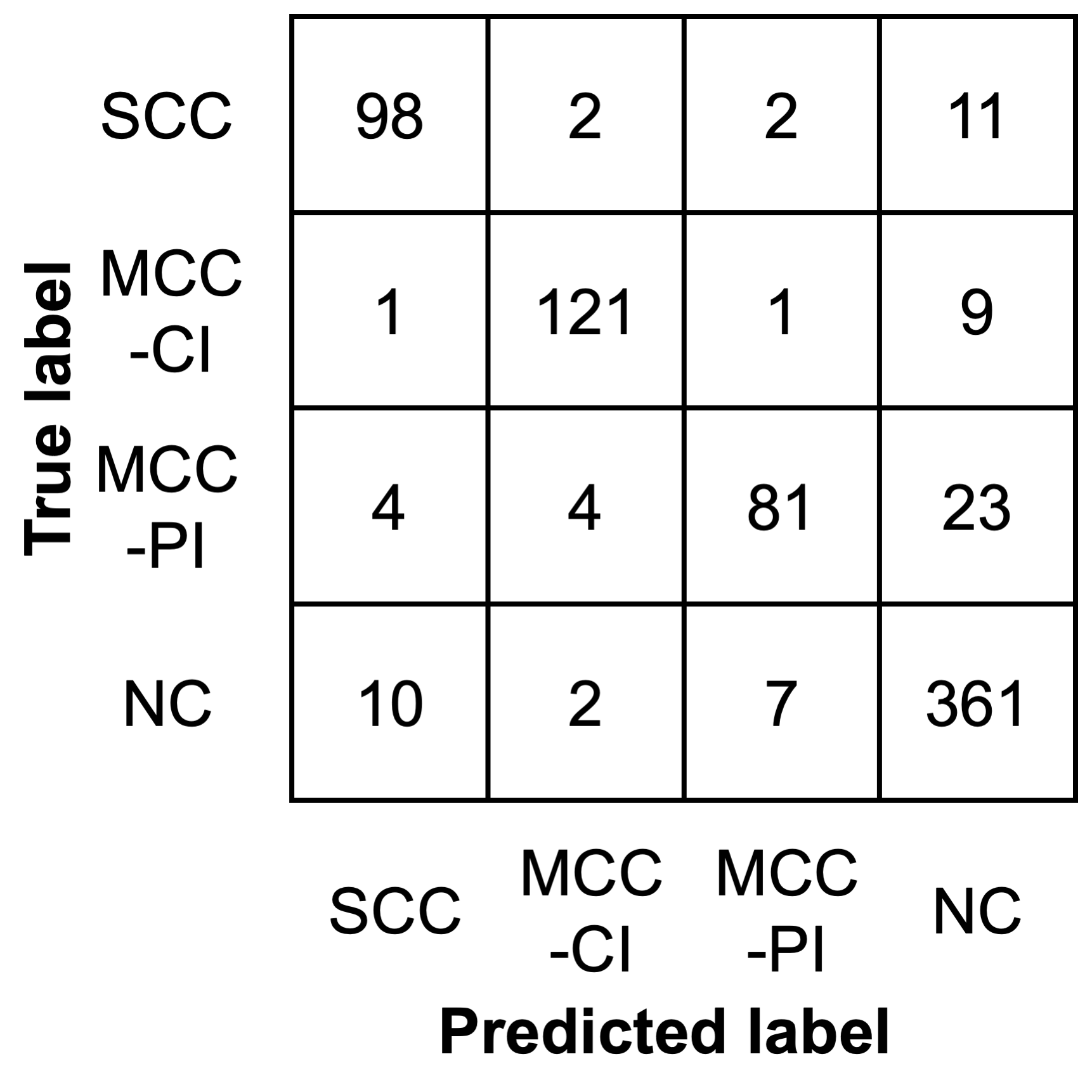}
    \vspace{-0.15in}
    \caption{Confusion matrix for all table types}
    \label{fig:confusion}
    \vspace{-0.30in}
\end{wrapfigure}

    
    
    


\section{Conclusions} 
\label{sec:conclusions}
\vspace{-0.15cm}
We define the novel and challenging task of extracting material compositions from tables in scientific papers. This task has importance beyond material science, since many other scientific disciplines use tables to express compositions in their domains. We harvest a dataset using distant supervision, combining information from a MatSci DB with tables in respective papers. We present a strong baseline system \sys, for this task. It encodes tables as graphs and trains GNNs for table-type classification. Further, to handle incomplete information in PI tables, it includes the text associated with the tables from respective papers. To handle domain-specific regular languages, a 
rule-based composition parser helps the model by extracting chemical compounds, numbers, units, and composition expressions. We find that our \sys{} baseline 
outperforms other architectures that linearize the tables by huge margins. In the future, our work can be extended to extract material properties that are also often found in tables. The code and data are made available in the \href{https://github.com/M3RG-IITD/DiSCoMaT}{GitHub repository} of this work.

\section*{Acknowledgements}
N. M. Anoop Krishnan acknowledges the funding support received from SERB (ECR/2018/002228), DST (DST/INSPIRE/04/2016/002774), BRNS YSRA (53/20/01/2021-BRNS), ISRO RESPOND as part of the STC at IIT Delhi. Mohd Zaki acknowledges the funding received from the PMRF award by Government of India. Mausam acknowledges grants by Google, IBM, Verisk, and a Jai Gupta chair fellowship. He also acknowledges travel support from Google and Yardi School of AI travel grants. The authors thank the High Performance Computing (HPC) facility at IIT Delhi for computational and storage resources. 

\section*{Limitations and outlook} \sys\ is a pipelined solution trained component-wise. This raises a research question: can we train one end-to-end trained ML model that not only analyzes a wide variety of table structures but also combines the understanding of regular expressions, extraction of chemical compounds and scientific units, textual understanding and some mathematical processing? This defines a challenging ML research question and one that can have a direct impact on the scientific MatSci community. \\
Indeed, automating parts of scientific discovery through such NLP-based approaches has the potential for biases and errors. Note that wrong and biased results can lead to erroneous information about materials. To a great extent, this issue is addressed as we rely only on published literature. The issue could be further addressed by considering larger datasets covering a wider range of materials.

\bibliography{references}

\appendix

\newpage
\section{Appendix} 
\label{sec:appendix}

\begin{figure}[hbt]
    \centering
    \includegraphics[width=0.35\textwidth, height=2.5cm]{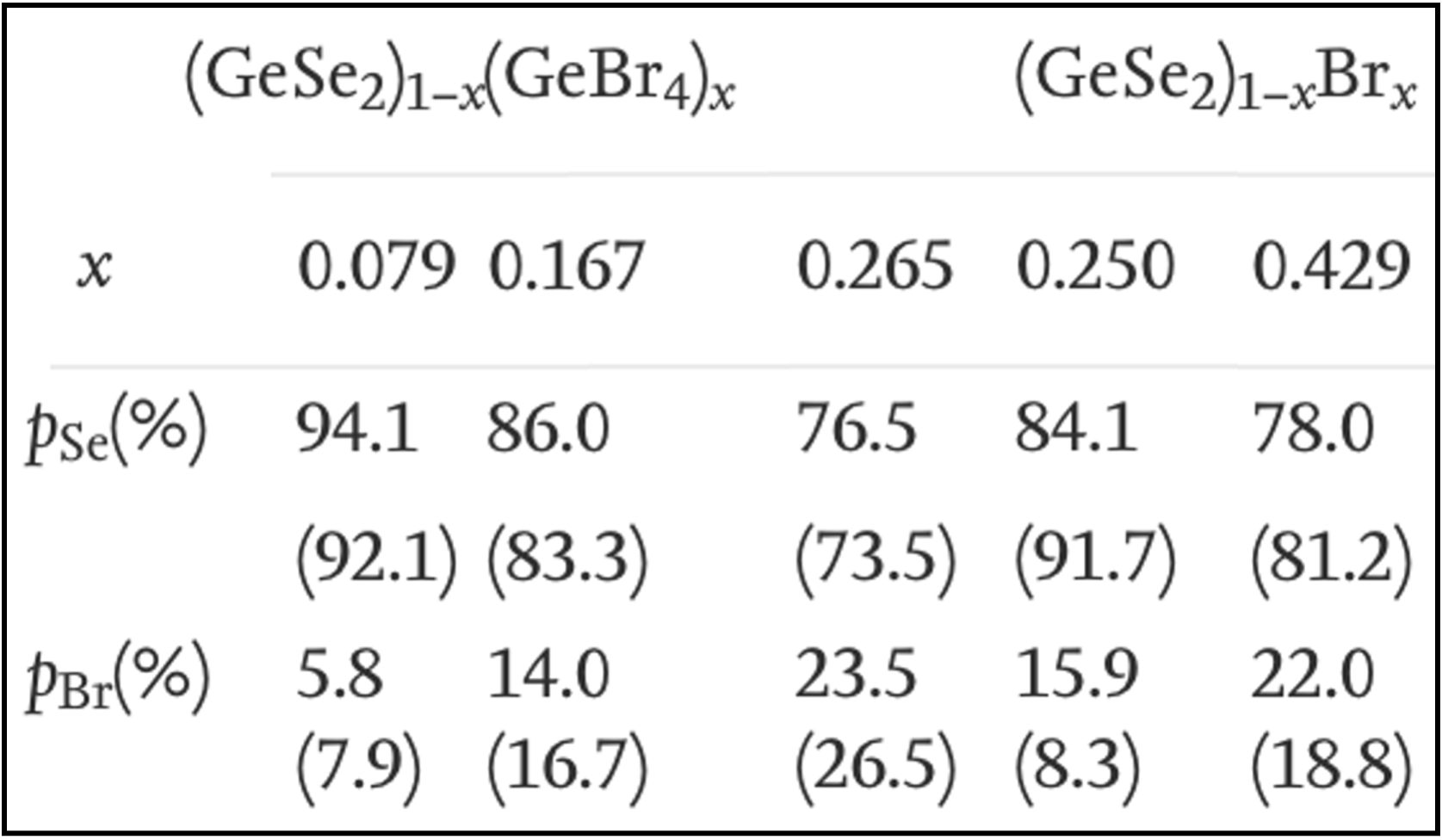}
    \caption{Percentages as variables \cite{fig_2_regex}}
    \label{fig:regex_var_comp_table}
    \vspace{-2ex}
\end{figure}


\subsection{Constraint-aware training}
\label{app:constraint}
As discussed in Section \ref{sec:constraints}, to encourage GNN$_2$ to make structurally consistent predictions, we express a set of constraints on the complete labeling as follows. (1) A row and a column cannot both have compositions or constituents. (2) Composition and material ID must be orthogonally predicted (i.e., if a row has a composition, then the ID must be predicted in some column, and vice versa). (3) Constituents and material IDs must never be orthogonally predicted (that is, if rows have constituents, then another row in the table must have the ID). And, (4) material ID must occur at most once for the entire table.
Let $r_i$ and $c_j$ be the predicted labels of row $i$ and column $j$. Further, let $\theta$ represent GNN$_2$'s parameters.

Constraint (1) is expressed as a hard constraint by:
\begin{center}
\(
r_i = l \Rightarrow c_j \neq l \) \(~~~~\forall i\in\{1,\ldots, R\}, j\in \{1,\ldots,C\},\; l\in\{1,2\}.
\) \end{center}
The equivalent probabilistic statement is:
\begin{center}
\(
P(r_i = l; \theta) + P(c_j = l; \theta) - 1 \leq 0\) \( \forall i \in \{1, \dots, R\}, \; j \in \{1, \dots, C\}, \; l \in \{1, 2\}.
\) \end{center}

Constraint (2) can be written in the form of hard constraints as:
\begin{center}
\(
r_{i_1} = 1 \Rightarrow r_{i_2} \neq 3 ~~\forall i_1, i_2 \in \{1,\ldots, R\}, i_1 \neq i_2.
\) \end{center}
\begin{center}
\(
c_{j_1} = 1 \Rightarrow c_{j_2} \neq 3 ~\forall j_1, j_2 \in \{1,\ldots, C\}, j_1 \neq j_2.
\) \end{center}
Equivalent probabilistic statements are:
\begin{center}
\(
P(r_{i_1} = 1; \theta) + P(r_{i_2} = 3; \theta) - 1 \leq 0\) \( \forall i_1,\;i_2 \in \{1, \ldots, R\}, \; i_1 \neq i_2.
\) \end{center}
\begin{center}
\(
P(c_{j_1} = 1; \theta) + P(c_{j_2} = 3; \theta) - 1 \leq 0\) \( \forall j_1,\;j_2 \in \{1, \ldots, C\}, \; j_1 \neq j_2.
\) \end{center}

We write constraint (3) in a hard constraint form as:
\begin{center}
\(
r_i = l \Rightarrow c_j \neq 5-l\) \( \forall i\in\{1,\ldots, R\},\; j\in \{1,\ldots,C\},\; l\in\{2,3\}.
\) \end{center}
The equivalent probabilistic statement is:
\begin{center}
\(
P(r_i = l; \theta) + P(c_j = 5-l; \theta) - 1 \leq 0\) \( \forall i \in \{1, \dots, R\}, \; j \in \{1, \dots, C\}, \; l \in \{2, 3\}.
\) \end{center}

Finally, hard versions of constraint (4) can be stated as:
\begin{center}
\(
r_{i_1} = 3 \Rightarrow r_{i_2} \neq 3 ~~~~ 1 \leq i_1 < i_2 \leq R.
\) \end{center}
\begin{center}
\(
c_{j_1} = 3 \Rightarrow c_{j_2} \neq 3 ~~~~ 1 \leq j_1 < j_2 \leq C.
\) \end{center}
\begin{center}
\(
r_i = 3 \Rightarrow c_j \neq 3 ~~~~ \forall i \in \{1, \dots, R\}, \; j \in \{1, \dots, C\}.
\) \end{center}
Equivalent probabilistic statements are:
\begin{center}
\(
P(r_{i_1} = 3; \theta) + P(r_{i_2} = 3; \theta) - 1 \leq 0\) \( 1 \leq i_1 < i_2 \leq R.
\) \end{center}
\begin{center}
\(
P(c_{j_1} = 3; \theta) + P(c_{j_2} = 3; \theta) - 1 \leq 0\) \( 1 \leq j_1 < j_2 \leq C.
\) \end{center}
\begin{center}
\(
P(r_i = 3; \theta) + P(c_j = 3; \theta) - 1 \leq 0\) \(\forall i \in \{1, \dots, R\}, \; j \in \{1, \dots, C\}.
\) \end{center}
As explained in Section \ref{sec:constraints}, we convert all these probabilistic statements to an auxiliary penalty term, which gets added to the loss function.

\subsection{Dataset details}
\label{app:dataset-details}

We use the INTERGLAD V7.0 (Interglad) database \cite{interglad} for annotating our training set as described in Section \ref{sec:experiments}. Since the Interglad database is not publicly available, we use SciGlass \cite{sciglass} database (released under Open Database License) as a proxy for Interglad in the shared code. Interglad contains 12634 compositions corresponding to the publications in our training set. However, SciGlass contains only 2347 compositions of these publications.
Hence, the code provided by us can annotate a subset of the training data only.
However, we do provide training data annotated using the Interglad database for reproducing the results of \sys{} for training and evaluation.
Also, anyone with Elsevier and Interglad subscriptions can replicate our training set (by replacing SciGlass database files with Interglad database files).



\begin{figure*}[!htb]
\begin{center}
    \centering
    \includegraphics[width=\textwidth]{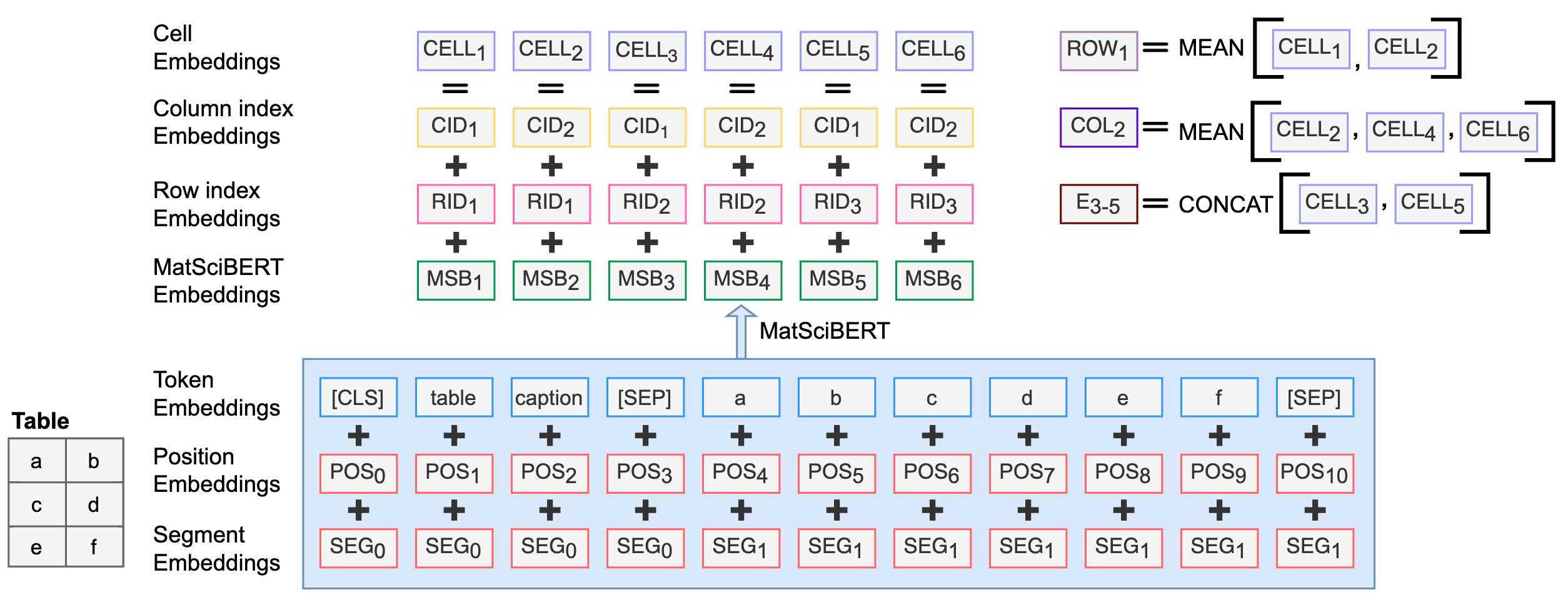}
    \caption{Schematic of \textsc{TaPas-Adapted} baseline model}
    \label{fig:tapas_adapted}
\end{center}
\end{figure*}

\begin{table}
\begin{subtable}[c]{0.49\textwidth}
\centering
\resizebox{0.6\textwidth}{!}{
\begin{tabular}{lccc}
    \toprule &
    \multicolumn{3}{c}{ \textbf{Splits}}\\
    \cmidrule(lr){2-4}
    \textbf{Table Type} &  \textbf{Train} & \textbf{Dev} & \textbf{Test} \\
    \midrule
    SCC & 704 & 110 & 113 \\
    MCC-CI & 626 & 132 & 132 \\
    MCC-PI & 317 & 109 & 112 \\
    NC & 2761 & 387 & 380 \\
    \midrule
    Total & 4408 & 738 & 737 \\
    \bottomrule
\end{tabular}}
\subcaption{}
\label{tab:dataset_stats_1}
\end{subtable}
\begin{subtable}[c]{0.49\textwidth}
\centering
\resizebox{0.6\textwidth}{!}{
\begin{tabular}{lccc}
    \toprule &
    \multicolumn{3}{c}{ \textbf{Splits}}\\
    \cmidrule(lr){2-4}
     & \textbf{Train} & \textbf{Dev} & \textbf{Test} \\
    \midrule
    Publications & 1880 & 330 & 326 \\
    Materials & 11207 & 2873 & 2649 \\
    Tuples & 38799 & 10168 & 9514 \\
    \bottomrule
\end{tabular}}
\subcaption{}
\label{tab:dataset_stats_2}
\end{subtable}
\caption{Number of (a) each of the table types and (b) journals from which the tables are obtained, materials in the tables, and the tuples for the three splits.}
\label{tab:dataset_stats}
\end{table}

We have manually annotated the val and test set, due to the fact that distantly supervised annotations can have noise and are not always 100\% accurate. The inter-annotator agreement has already been discussed in \ref{sec:dataset_construction}. Along with the provision of manual annotation, the in-house annotation tools also contained several checks on conditions that shouldn't arise such as: whether the annotator has missed annotating any table, or the annotator has annotated with out-of-range labels or a row/column having both composition and constituent or vice-versa i.e. composition/constituent present in both row and column of a table. With the help of these self-checks and mutual discussions on disagreements, we annotated our val and test dataset.

Table \ref{tab:dataset_stats} presents some statistics about our dataset. Table \ref{tab:dataset_stats_1} shows the number of tables in our dataset belonging to different table types. Further, Table \ref{tab:dataset_stats_2} shows the total number of publications, materials, and tuples in all three splits. We release our code and data under the Creative Commons Attribution-NonCommercial-ShareAlike 4.0 (CC BY-NC-SA 4.0) International Public License.

\begin{figure*}[htb]
\begin{center}
    \centering
    \begin{subfigure}[b]{0.32\textwidth}
        \centering
        \caption{}
        \includegraphics[width=\textwidth]{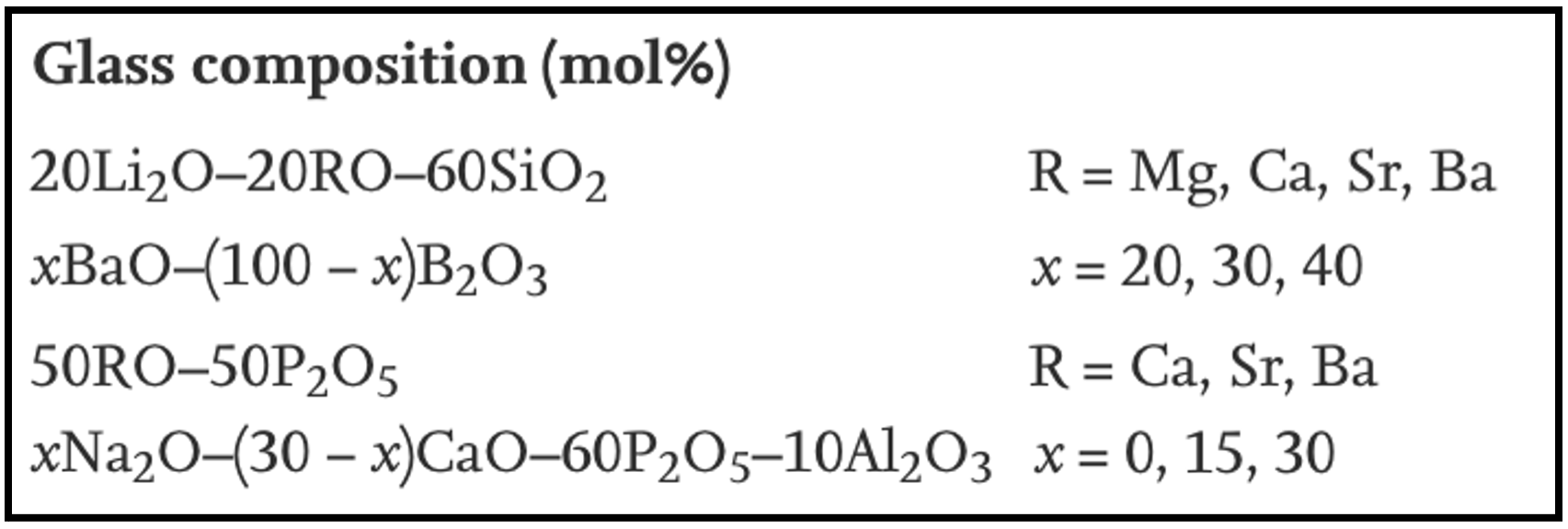}
        \label{fig:corner_scc_1}
    \end{subfigure}
    \hfill
    \begin{subfigure}[b]{0.32\textwidth}
        \centering
        \caption{}
        \includegraphics[width=\textwidth]{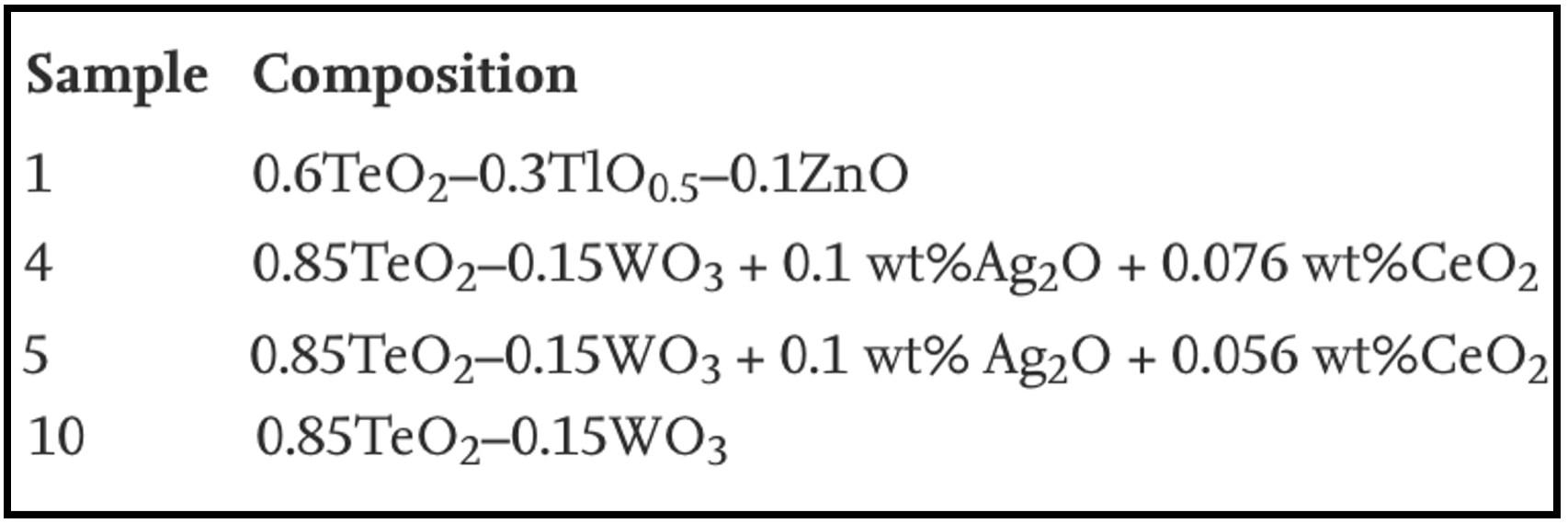}
        \label{fig:corner_scc_2}
    \end{subfigure}
    \hfill
    \begin{subfigure}[b]{0.32\textwidth}
        \centering
        \caption{}
        \includegraphics[width=\textwidth]{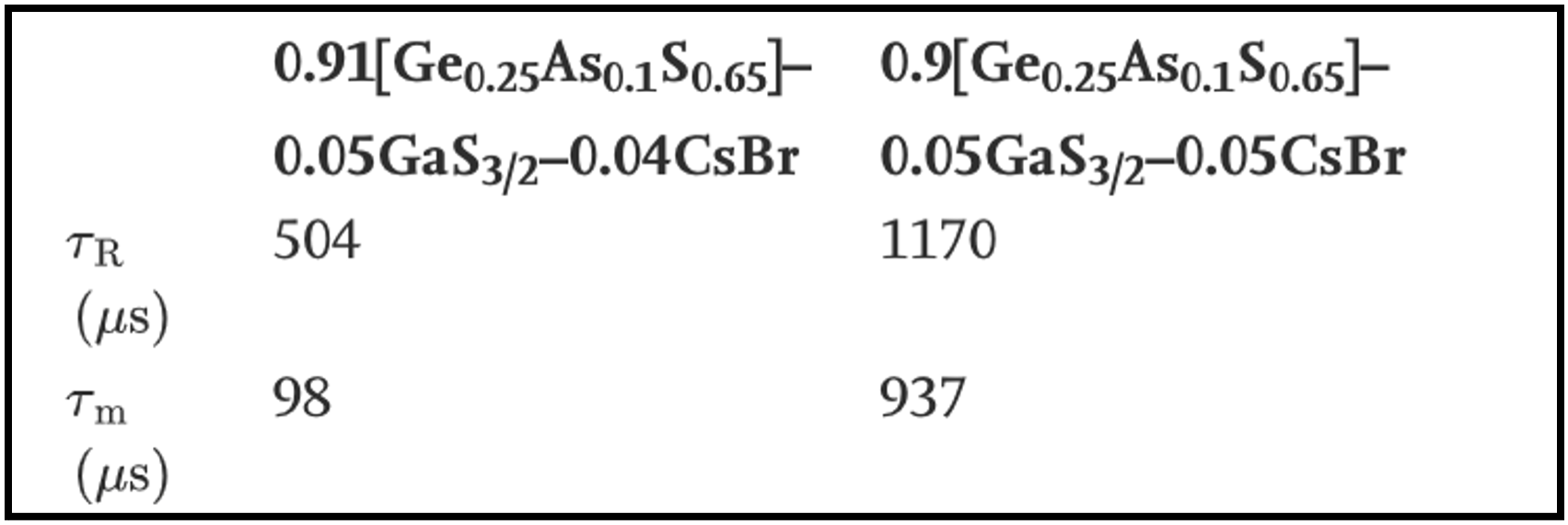}
        \label{fig:corner_scc_3}
    \end{subfigure}
    \vspace{-0.3cm}
    \caption{Examples of corner case composition tables (a) \cite{scc_1_S0022309304010804} (b) \cite{scc_2_S0022309309005109} (c) \cite{scc_3_S0022309302009468}}
    \label{fig:corner_scc}
    \vspace{-0.2cm}
\end{center}
\end{figure*}
\begin{figure*}[htb]
\begin{center}
    \centering
    \begin{subfigure}[b]{0.32\textwidth}
        \centering
        \caption{}
        \includegraphics[width=\textwidth]{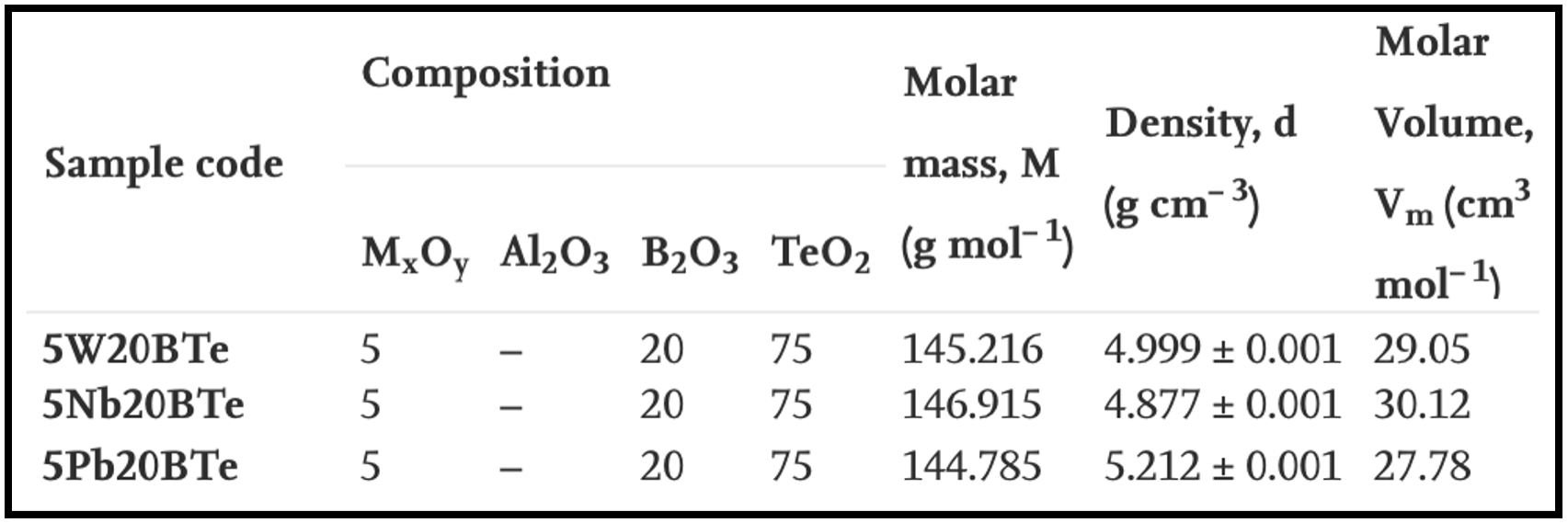}
        \label{fig:corner_mcc_1}
    \end{subfigure}
    \hfill
    \begin{subfigure}[b]{0.32\textwidth}
        \centering
        \caption{}
        \includegraphics[width=\textwidth]{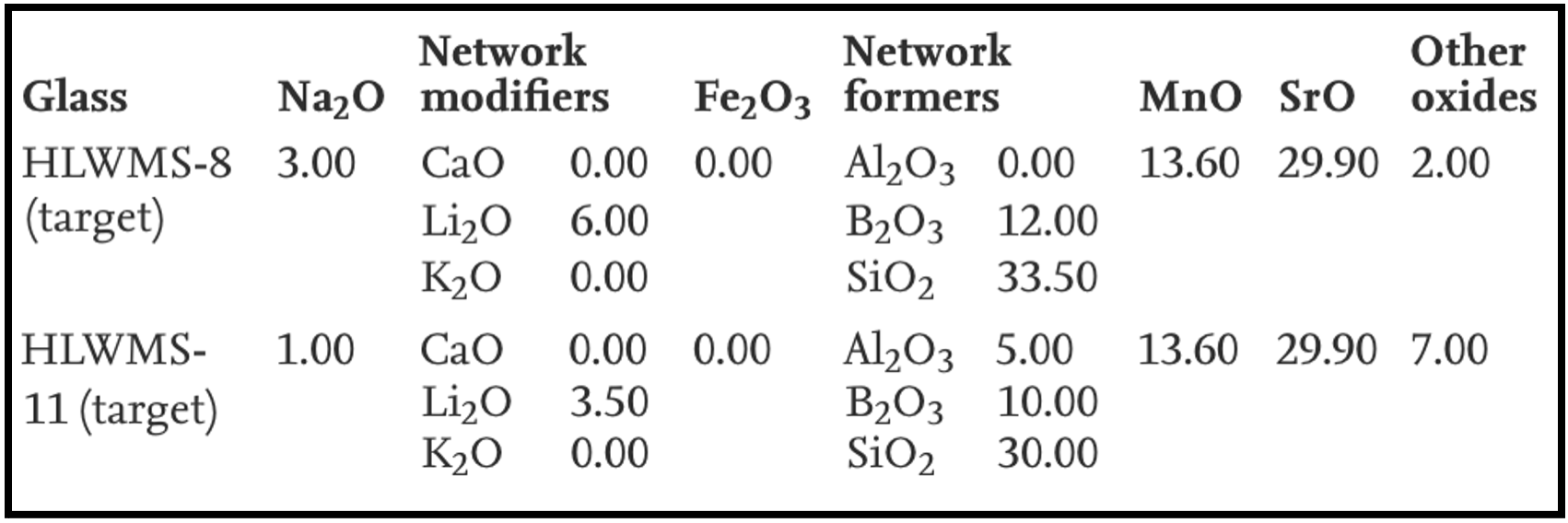}
        \label{fig:corner_mcc_2}
    \end{subfigure}
    \hfill
    \begin{subfigure}[b]{0.32\textwidth}
        \centering
        \caption{}
        \includegraphics[width=\textwidth]{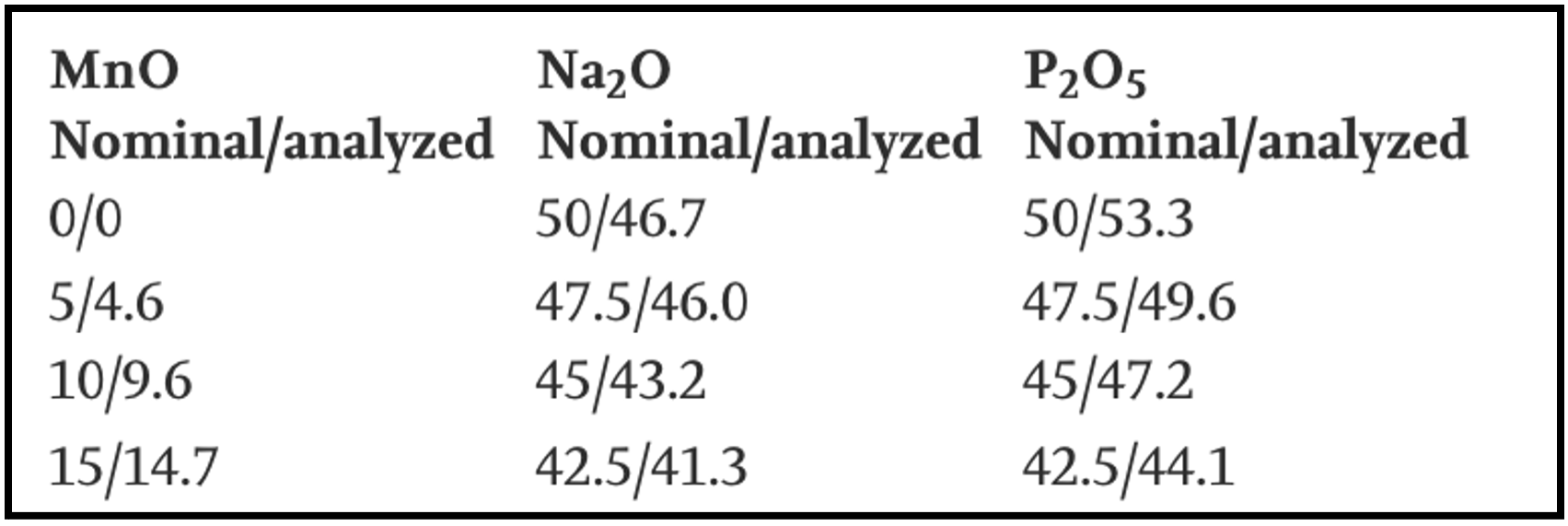}
        \label{fig:corner_mcc_3}
    \end{subfigure}
    \vspace{-0.3cm}
    \caption{Some more examples of corner case composition tables (a) \cite{mcc_ci_1_S0022309315301824} (b) \cite{mcc_ci_2_S0022309302018161} (c) \cite{mcc_ci_3_S0022309314000957}}
    \label{fig:corner_mcc}
    \vspace{-0.2cm}
\end{center}
\end{figure*}

\subsection{Baseline models}
\label{app:baseline}
In this section, we describe the details of our baseline models: \textsc{TaPas}, \textsc{TaPas-Adapted}, \textsc{TaBERT} and \textsc{TaBERT-Adapted}. Since, the \textsc{TaPas} \cite{herzig-etal-2020-tapas} architecture has been used for QA over tables and we do not have any questions in the composition extraction task, we use table caption as a proxy for the question. We replace the empty table cells with a special [EMPTY] token. The table caption and text in table cells are converted to word-pieces using the $LM$ tokenizer. Then, we concatenate the word-pieces of the caption and row-wise flattened table. Note that it is possible to obtain more than one word-piece for some table cells. Since the input length after tokenization can be greater than 512, we truncate the minimum possible rows from the end so that the length becomes less than or equal to 512. To avoid a large number of rows getting truncated due to long captions, we truncate the caption so that it only contributes $\leq$ 100 word-pieces. To differentiate between the table cells belonging to different rows and/or columns, row and column index embeddings are added to the word-piece embeddings in the \textsc{TaPas} architecture. Position and Segment embeddings are the same as in BERT \cite{devlin2018bert}, except that position indexes are incremented when the table cell changes. Original \textsc{TaPas} architecture also involves adding different Rank embeddings to the input in order to answer rank-based questions. We use the same rank embeddings for every table cell since there is no rank relation among the table cells for our case. \\
All these different types of embeddings are added together and passed through the $LM$. We take the contextual embedding of the first word-piece of every table cell to be representative of it. Since we do not have row and column nodes here, row and column embeddings are computed by taking the average of the first word-piece contextual embeddings of cells occurring in that row/column, which are then fed to an MLP for row/column classification. Edge embeddings are computed by concatenating the first workpiece contextual embeddings of source and destination cells. \\
Figure \ref{fig:tapas_adapted} shows the schematic of \textsc{TaPas-Adapted} model. Here, we initialize $LM$ weights with that of \textsc{MatSciBERT} \cite{gupta_matscibert_2022}. All other details are the same as in the \textsc{TaPas} model, except that here we add row and column index embeddings to \textsc{MatSciBERT} output, instead of input. \\
For \textsc{TaBERT} also, we use the table caption as the proxy for the NL sentence, concatenate it with linearized rows and feed into the \textsc{TaBERT} model which generates cell embedding by passing through BERT and applying vertical attention to propagate information across columns. Following the kind of linearization used by \textsc{TaBERT}, we linearize each cell as a concatenation of cell type and cell value for each cell, where cell type is divided into numeric, alphanumeric or text. Since \sys\; does not use pretraining, we do not use \textsc{TaBERT's} pretrained weights but instead train from initial weights on our row, column and edge-level prediction tasks. We also implement another baseline called \textsc{TaBERT-Adapted}, which replaces the \textsc{BERT} encoder in \textsc{TaBERT} with \textsc{MatSciBERT} \cite{gupta_matscibert_2022} to provide materials science domain's information to the model.\\
In \textsc{TABBIE}, as opposed to \textsc{TaPaS} and \textsc{TaBERT}, table cells are passed independently into the LM, instead of being linearized/flattened into a single long sequence. Similar to \textsc{TaBERT}, we don't initialize \textsc{TABBIE}'s architecture with its pretrained weights for a fair comparison. \textsc{TABBIE-Adapted} again replaces the \textsc{BERT} encoder in \textsc{TaBERT} with \textsc{MatSciBERT} \cite{gupta_matscibert_2022}.
\\
The complete code and data is available at \href{https://github.com/M3RG-IITD/DiSCoMaT}{https://github.com/M3RG-IITD/DiSCoMaT}.

\subsection{Implementation details}
\label{app:implementation}
For Graph Attention Networks (GATs) \cite{velickovic2018graph}, we use the GAT implementation of Deep Graph Library \cite{dgl_citation_wang2019}. For LMs, \textsc{TaPas}, we use the implementation by Transformers library \cite{transformers_huggingface}. We use \textsc{TaBERT}'s source code from their GitHub repository. We implement and train all models using PyTorch  \cite{pytorch_NEURIPS2019_9015} and AllenNLP \cite{Gardner2017AllenNLP}. We optimize the model parameters using Adam \cite{adam_opt_kingma} and a triangular learning rate \cite{triangle_lr}. We further use different learning rates for $LM$ and non-LM parameters (GNNs, MLPs) (App.~\ref{app:hyperparameter}). To deal with imbalanced labels, we scale loss for all labels by weights inversely proportional to their frequency in the training set. All experiments were run on a machine with one 32 GB V100 GPU. Each model is run with three seeds and the mean and std. deviation is reported.


\subsection{Hyper-parameter details}
\label{app:hyperparameter}
Now, we describe the hyper-parameters of \sys. Both GNN$_1$ and GNN$_2$ can have multiple hidden layers with different numbers of attention heads. We experiment with hidden layer sizes of 256, 128, and 64 and the number of attention heads as 6, 4, and 2. We include residual connections in GAT, exponential linear unit (ELU) non-linearity after hidden layers, and LeakyRELU non-linearity (with slope $\alpha$ = 0.2) to compute attention weights as done in \cite{velickovic2018graph}. Training is performed using 8 tables in a batch and we select the checkpoint with the maximum dev MatL F$_1$ score. We use a triangular learning rate and choose the peak learning rate for $LM$ to be among 1e-5, 2e-5, and 3e-5 and the peak learning rate for non-$LM$ parameters to be among 3e-4 and 1e-3. A warmup ratio of 0.1 is used for all parameters. We further use batch normalization \cite{batch_norm} and dropout \cite{dropout} probability of 0.2 in all MLPs. We use the same $\lambda$ for every constraint penalty term. Embedding sizes for features are chosen from 128 and 256 and edge loss weight is selected among 0.3 and 1.0.

\begin{table}[htb]
\centering
\resizebox{0.48\textwidth}{!}{
\begin{tabular}{lcc}
    \toprule
    \textbf{Hyper-parameter} & \textbf{GNN$_1$} & \textbf{GNN$_2$} \\
    \midrule
    GAT Hidden Layer Sizes & [256, 128, 64] & [128, 128, 64] \\
    GAT Attention Heads & [4, 4, 4] & [6, 4, 4] \\
    Peak LR for $LM$ & 1e-5 & 2e-5 \\
    Peak LR for non-$LM$ & 3e-4 & 3e-4 \\
    RegEx feature emb size & 256 & NA \\
    Max-frequency feature emb size & 256 & 128 \\
    Constraint penalty ($\lambda$) & 50.0 & 30.0 \\
    Edge loss weight & NA & 1.0 \\
    \bottomrule
    \end{tabular}}
\caption{Hyper-parameters for \sys.}
\label{tab:hyperparams}
\end{table}


\subsection{Corner cases}
\label{app:corner}

Figure \ref{fig:corner_scc} shows examples of some corner case tables. In Figure \ref{fig:corner_scc_1}, elements are being used as variables. Moreover, the values that variables can take are present in a single cell only. Figure \ref{fig:corner_scc_2} shows a table where units occur within the composition itself. Also, mixed units are being used to express the composition. Figure \ref{fig:corner_scc_3} comprises compositions having both elements and compounds. Whereas, we made different REs for element compositions and different REs for compound compositions. Hence our REs are unable to match these.

Figure \ref{fig:corner_mcc} shows some more examples of corner cases. In Figure \ref{fig:corner_mcc_1}, the first compound has to be inferred using the Material IDs. For example, W corresponds to WO$_3$ and Nb corresponds to Nb$_2$O$_5$. \sys{} makes the assumption that composition is present in a single row/column. Figure \ref{fig:corner_mcc_2} refutes this assumption as compositions are present in multiple rows. Sometimes researchers report both theoretical (nominal) and experimental (analyzed) compositions for the same material. The table in Figure \ref{fig:corner_mcc_3} lists both types of compositions in the same cell and hence can't be extracted using \sys{}.

\newpage
\onecolumn

\end{document}